\documentclass[lettersize,journal,twocolumn]{IEEEtran}
\usepackage[top=57pt, bottom=43pt, left=48pt, right=48pt]{geometry}
\usepackage{amsmath,amsfonts}
\usepackage{algorithmic}
\usepackage{algorithm}
\usepackage{array}
\usepackage[caption=false,font=normalsize,labelfont=sf,textfont=sf]{subfig}
\usepackage{textcomp}
\usepackage{stfloats}
\usepackage{url}
\usepackage{verbatim}
\usepackage{graphicx}
\usepackage{xcolor}
\usepackage{cite}
\usepackage{hyperref}
\usepackage{multirow}
\usepackage{bbding}
\usepackage{booktabs}
\usepackage{adjustbox}
\usepackage{threeparttable}
\usepackage{algorithm, algorithmic}
\usepackage{makecell}
\title{RCGDet3D: Rethinking 4D Radar-Camera Fusion-based 3D Object Detection with Enhanced Radar Feature Encoding}

\author{Weiyi Xiong, and Bing Zhu
\thanks{Weiyi Xiong and Bing Zhu are with the School of Automation Science and Electrical Engineering, Beihang University, Beijing, P.R.~China
(e-mail: weiyixiong@buaa.edu.cn; zhubing@buaa.edu.cn).}
\thanks{Corresponding Author: Bing Zhu.}
}

\begin{document}

\maketitle

\begin{abstract}
4D automotive radar is indispensable for autonomous driving due to its low cost and robustness, yet its point cloud sparsity challenges 3D object detection. 
Existing 4D radar-camera fusion methods focus on complex fusion strategies, trading inference speed for marginal gains. 
This trade-off hinders real-time deployment due to heavy computation on dense feature maps.
In contrast, feature extraction from sparse radar points is less time-consuming but remains under-explored. 
This work uncovers that simply enhancing radar feature extraction can achieve comparable or even higher performance than elaborate fusion modules, while maintaining real-time performance.
Based on this finding, we propose RCGDet3D, which centers on radar feature encoding and simplifies multi-modal fusion. 
Its encoder inherits from the efficient Gaussian Splatting-based Point Gaussian Encoder (PGE) in RadarGaussianDet3D with two key improvements.
First, the Ray-centric PGE (R-PGE) predicts Gaussian attributes in ray-aligned coordinate systems before unifying them to Bird's-Eye View (BEV) space, significantly improving geometric consistency and reducing learning difficulty by decoupling the coordinate transformation from representation learning.
Second, a Semantic Injection (SI) module incorporates visual cues from images, producing more geometrically accurate and semantically enriched radar features.
Experiments on View-of-Delft (VoD) and TJ4DRadSet show that RCGDet3D outperforms state-of-the-art methods in both accuracy and speed, setting a new benchmark for real-time deployment.
\end{abstract}

\section{Introduction}\label{sec:intro}

As an affordable alternative to LiDAR, 4D radar has drawn researchers' wide attention in the field of autonomous driving. 
Although 4D radar and LiDAR data are both typically in the form of 3D point clouds, there are differences between them, among which point density is the most obvious. 
For example, in the View-of-Delft (VoD) dataset \cite{VoD}, the average number of 4D radar points in the camera field of view (FoV) in a scan is about 210, while for LiDAR it is 24,000. 
Nevertheless, most 4D radar-based models follow LiDAR-based ones to adopt the voxel/pillar encoder to extract the radar voxel/BEV features, resulting in sparse feature maps.
Despite various of modifications are made, such as separating the positional, intensity and velocity features \cite{RCFusion}, replacing the pillar feature net into the attention mechanism \cite{RPFA-Net}, or additionally estimating the point density \cite{SMURF}, they do not change the essence of voxelization.

Recently, RadarGaussianDet3D \cite{RadarGaussianDet3D} proposes a Gaussian Splatting-based radar feature encoder called Point Gaussian Encoder (PGE), where each radar point is expanded into a Gaussian ellipsoid (i.e., Gaussian primitive) with corresponding scales and orientations, and Bird's-Eye View (BEV) rasterization is performed to obtain the radar BEV feature map.
As the Gaussian Splatting technique is famous for its high rendering quality and speed, the PGE can run even faster than the pillar encoder, generating denser and more accurate BEV features.
In addition, the quantization error in the pillarization/voxelization process can be reduced in PGE, as a point can contribute to multiple neighboring grids.

However, the predicted Gaussian primitives are directly splatted onto the BEV plane, i.e., they are expected to be represented in the perception coordinate (ego coordinate), making learning difficult as the model need to implicitly estimate the direction of rays and the pose of sensor when predicting Gaussian attributes. 
Some camera-based BEV perception methods have also noticed the similar phenomenon.
For example, GaussianBeV \cite{GaussianBeV} proposes to predict the Gaussian rotation relative to the optical ray, and by multiplying with the rotation quaternion from ray to the camera optical axis, it is transformed to camera coordinate system. 
Nevertheless, the issue is not fully addressed since its predicted positional offset is still in camera coordinate. 
In addition, it cannot adapt to the case when BEV augmentation is applied, because operations such as horizontal flipping cannot be represented into a quaternion. 
In this work, the predicted Gaussian primitives are completely represented in their respective ray-aligned coordinate systems, and are transformed into the ego coordinate according to point locations and the radar pose before BEV rasterization. 
By applying this modification, the predicted Gaussian primitives are called ray-centric Gaussian primitives, and the PGE evolves into Ray-centric PGE (R-PGE).

While the ray-centric Gaussian representation significantly improves geometric fidelity, radar point clouds alone still lack semantic cues, which are crucial for accurately predicting object boundaries and attributes. 
A straightforward solution is to incorporate visual information from cameras for compensation.
However, most works focus on enhancing the image view transformation with radar features \cite{LXL} or design complex fusion strategy \cite{MSSF,SGDet3D}.
While showing some improvements, these methods often introduce considerable latency due to the process of large dense feature maps, making them less practical for real-time applications. 
Moreover, the performance gain is constrained by the quality of extracted radar features. 
We find that replacing the commonly applied pillar encoder to R-PGE can bring a significant performance gain in 4D radar-camera fusion models, while maintaining similar inference speed.  
This result indicates that the performance bottleneck of 4D radar-camera fusion-based 3D object detection methods mainly lies in the radar feature extraction module, which is overlooked by most researchers. 

Based on the above discovery, we further propose a Semantic Injection (SI) sub-module, leveraging rich semantic information provided by the RGB images to assist the prediction of ray-centric Gaussian primitives when the camera modality is available. The introducing of image features not only enriches the Gaussian features, but also increases the accuracy of the physical attributes like scales and rotations. This simple enhancement further improves the detection performance with only minimal latency.

In this work, we introduce RCGDet3D by adding an image branch like LXL \cite{LXL} to RadarGaussianDet3D \cite{RadarGaussianDet3D}, and applying the aforementioned enhancements to the radar branch. In summary, our contributions are fourfold:
\begin{itemize}
    \item This work uncovers that the bottleneck of current 4D radar-camera fusion-based 3D object detectors is not the image view transformation or fusion strategy, but the under-explored radar feature representation. 
    \item Based on PGE in RadarGaussianDet3D, we propose Ray-centric Point Gaussian Encoder (R-PGE), to predict each Gaussian primitive in the corresponding ray-aligned coordinate system. This makes the model focus on predicting the Gaussian attributes without implicitly estimating the radar pose and ray direction, improving geometric consistency. 
    \item A Semantic Injection (SI) sub-module is optionally introduced in R-PGE to fuse visual features with radar features before predicting Gaussian primitives. It enriches semantic information and improves the accuracy of estimated Gaussian attributes.
    \item Experiments on View-of-Delft \cite{VoD} and TJ4DRadSet \cite{TJ4DRadSet} show that the proposed RCGDet3D significantly outperforms the state-of-the-art detectors in both accuracy and inference speed, setting a new benchmark for practical deployment.
\end{itemize}

The rest of the paper is organized as follows. Section \ref{sec:related work} discusses related work on 4D radar-based and 4D radar-camera fusion-based 3D object detection methods, as well as the application of 3D Gaussian Splatting on autonomous driving. Section \ref{sec:method} introduces the proposed model, and Section \ref{sec:experiments} exhibits and analyzes the experimental results. In the end, Section \ref{sec:conclusion} summarizes this work.

\section{Related Work}\label{sec:related work}
\subsection{4D Radar Point Cloud-based 3D Object Detection}\label{sec:work_radar}
Most 4D radar point cloud-based 3D object detectors follow LiDAR-based ones to apply a pillar-based encoder for its efficiency \cite{VoD}. However, due to the sparsity and noisiness of radar points, targeted enhancements are needed to extract informative features.

Earliest work propose to modify certain components of the pillar encoder. RCFusion \cite{RCFusion} separates the process of position, intensity and velocity information by applying three individual MLPs. RPFA-Net\cite{RPFA-Net} replaces the PointNet\cite{PointNet} with Pillar Feature Attention module to extract pillar features. 

Some work cascade additional modules or processes. 
For instance, RadarPillars\cite{RadarPillars} conduct self-attention for non-empty pillars after pillarization, while SD4R \cite{SD4R} densify the foreground radar points with virtual points before pillarization.

More researches add extra branches to enrich the radar BEV features. Parallel to the pillar encoder, RCBEVDet \cite{RCBEVDet} proposes to scatter a point feature to multiple BEV grids according to the radar cross section (RCS) value, while SMURF \cite{SMURF} predicts point density via Kernal Density Estimation (KDE). The current state-of-the-art model, MAFF-Net\cite{MAFF-Net} introduces an auxiliary clustering-based branch to address the noise and outliers in the radar modality. 

Despite these effective modifications, the pillar-based encoder is inherently suboptimal for 4D radar points, due to the existence of quantization error and feature sparsity. To mitigate these issues, RadarGaussianDet3D \cite{RadarGaussianDet3D} proposes Point Gaussian Encoder (PGE), where each point is expanded to a Gaussian primitive, and BEV Gaussian Splatting is adopted to obtain dense feature map. By fully replacing the pillar-based encoder, this Gaussian-based encoder not only improves the detection accuracy, but also runs faster, showing great potential for practical deployment.

\subsection{4D Radar-Camera Fusion-based 3D Object Detection}
As the radar point cloud is geometrically inaccurate and semantically insufficient, relying only the 4D radar modality is inadequate for achieving high-quality detection. 
Consequently, many researchers conduct multi-modal fusion to leverage the rich semantic from camera images for supplement. 

Existing efforts in this area can be broadly categorized into three main directions: radar feature extraction \cite{RCFusion, RCBEVDet}, image view transformation \cite{LXL}, and multi-modal feature fusion \cite{MSSF,SGDet3D,RaGS,SIFormer}.
Early work focus more on the first two directions. The first type rarely relies on image features, and has already been discussed in Section \ref{sec:work_radar}. A representative work of the second type is LXL \cite{LXL}, which adopts a depth-based sampling strategy to transform image perspective features to BEV. The view transformation process is also assisted by radar 3D occupancy predicted from the radar BEV feature map.
Recently, researches have shifted towards designing effective fusion modules. For example, MSSF \cite{MSSF} adopts a multi-stage sampling scheme to perform voxel-level fusion, while SGDet3D \cite{SGDet3D} proposes a Dual-branch Fusion and an Object-oriented Attention module to achieve cross-modal deep interaction.

Although these elaborate image view transformation and multi-modal fusion schemes improves detection accuracy, they suffer from two main limitations.
First, the processing of large dense feature maps or quantities of voxels requires a considerable computational overhead, hindering the models' real-time capability and limiting their deployment. 
For example, even on a powerful RTX 4090 GPU, the current state-of-the-art detector SIFormer \cite{SIFormer} and the most efficient one RaGS \cite{RaGS} only runs 6.9 and 10.5 FPS respectively, on the VoD \cite{VoD} dataset.
This leaves a significant gap for real-time deployment on resource-constrained embedded platforms commonly used in autonomous vehicles. 
Moreover, the effectiveness of these proposed strategies heavily relies on the quality of extracted radar features; when radar features are insufficient or noisy, the performance improvements become marginal. 
These observations suggest that revisiting the relatively fast yet under-explored radar feature extraction approaches is a promising direction for improving both accuracy and efficiency in 4D radar-camera fusion.

\begin{figure*}
    \centering
    \includegraphics[width=\linewidth]{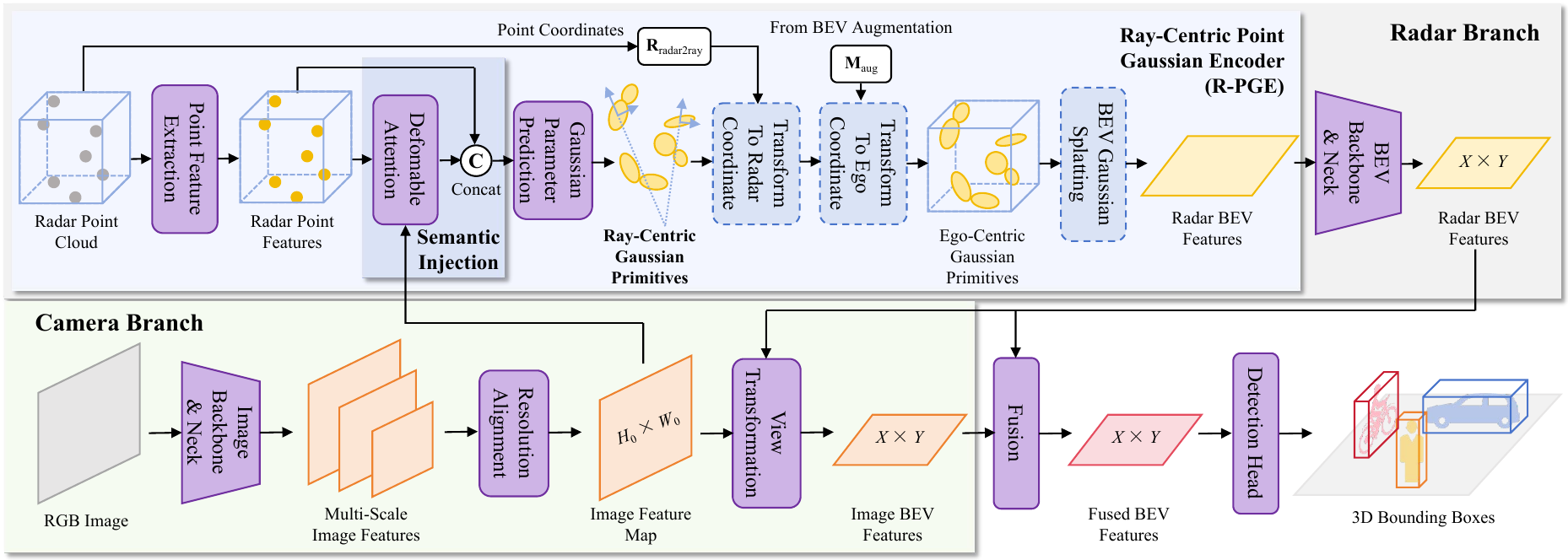}
    \caption{The overall architecture of RCGDet3D.
    Unlike existing methods that pursue sophisticated fusion strategies at the cost of speed, RCGDet3D demonstrates that enhancing radar feature extraction is the key to both high accuracy and high efficiency. 
    The Ray-centric Point Gaussian Encoder (R-PGE) predicts ray-centric Gaussian primitives from radar points, and employs Gaussian Splatting to generate radar BEV feature map.
    In addition, a Semantic Injection (SI) sub-module incorporates visual cues from the camera modality, yielding more precise Gaussian attributes and enriched point features. 
    For efficiency, a simplified image branch from LXL \cite{LXL}, as well as a light-weighted concatenation-based fusion is adopted.}
    \label{fig:overview}
\end{figure*}

\subsection{Gaussian Splatting in Autonomous Driving Perception}

Gaussian Splatting \cite{3DGS} has attracted widespread attention in the field of vision-based novel view synthesis (NVS) and 3D reconstruction, owing to its sparse yet expressive scene representation and high-quality real-time rendering. 
Recently, scholars have begun to explore its potential for camera-based perception tasks.

Since perception models require strong generalization capability, a feed-forward prediction paradigm is needed, instead of per-scene optimization in NVS. 
In these work, the Gaussian parameters can be either predicted from extracted features in one shot, or iteratively refined from queries. 
Single-shot prediction approaches typically generate each Gaussian primitive from a pixel of the 2D feature map \cite{GaussianBeV, GaussianLSS}, or from a voxel in the voxel space after feature lifting \cite{GSRender}. 
Iterative refinement methods, on the other hand, initialize a set of learnable queries and update them through deformable attention with image features, producing final Gaussian primitives \cite{GaussianFormer, ODG}.

In Gaussian Splatting-based NVS models, the Gaussian attributes include 3D mean, 3D scales, rotation, opacity and color \cite{3DGS}. However, from NVS to perception tasks, the goal shifts from appearance reconstruction to higher-level scene understanding. As a result, feature vectors \cite{GaussianBeV} or semantic logits \cite{GaussianFormer} instead of the color attribute are estimated. In addition, certain attributes can also be omitted for simplification or to facilitate training. For example, GSRender \cite{GSRender} does not predict Gaussian rotation, while GaussianLSS \cite{GaussianLSS} aligns one of the principal axes of each Gaussian ellipsoid with the corresponding pixel ray, and estimates the scale only along that axis. Finally, the Gaussian primitives are rasterized into BEV feature map or 3D semantic occupancy.

Despite the growing interest, the vast majority of related researches focus on camera inputs and dense prediction tasks like BEV segmentation and 3D occupancy prediction. Among the very few 4D radar-based 3D object detection methods, RaGS \cite{RaGS} adopts the query-based Gaussian generation strategy, where queries are initialized with the assistance of radar points. On the other hand, RadarGaussianDet3D \cite{RadarGaussianDet3D} transforms each radar point to a Gaussian primitive, and obtains the radar BEV feature map via feature rendering.

\section{Proposed Method}\label{sec:method}
\subsection{Overview}
Fig. \ref{fig:overview} illustrates the overall framework of the proposed RCGDet3D. The input radar point cloud and image are sent to modality-specific branches to extract corresponding BEV feature map, and the detection head predicts the 3D bounding boxes from the fused BEV features.

\begin{figure}
    \centering
    \includegraphics[width=\linewidth]{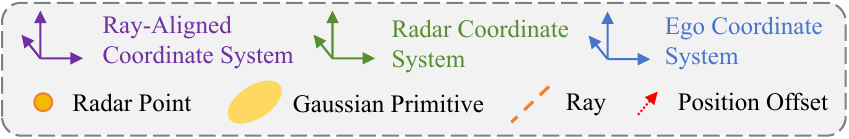}\\\vspace{2mm}
    \includegraphics[width=\linewidth]{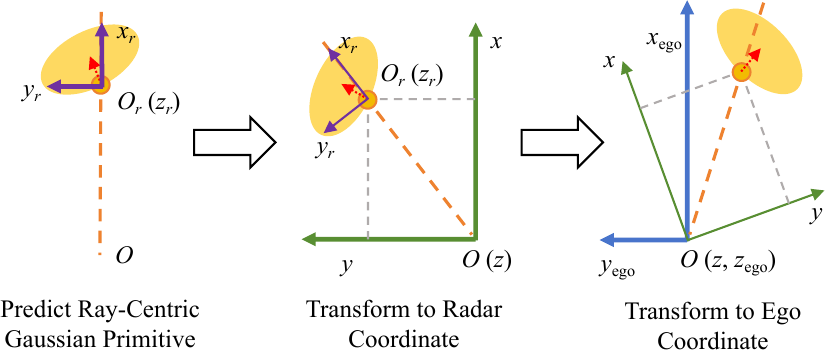}
    \caption{The coordinate transformation of ray-centric Gaussian Primitives (top-down view). }
    \label{fig:gaussian_coord_trans}
\end{figure}

The radar branch is based on RadarGaussianDet3D \cite{RadarGaussianDet3D} for its effectiveness and low latency. 
It is composed of a Point Gaussian Encoder (PGE) and the BEV backbone \& neck. 
The PGE predicts a Gaussian primitive for each radar point according to the extracted point features, and subsequently adopts BEV Gaussian Splatting to obtain the radar BEV feature map. 
However, as discussed in Section \ref{sec:intro}, the original PGE predicts ego-centric Gaussian primitives, raising difficulty for model learning. 
In our RCGDet3D, we propose Ray-centric PGE (R-PGE), where ray-centric Gaussian primitives are predicted instead of ego-centric ones to facilitate training. 
The Gaussian primitives are represented in their respective ray-aligned coordinate systems rather than the unified ego coordinate. 
Before splatted onto the BEV plane, the predicted Gaussian primitives are transformed into radar coordinate according to the azimuth and elevation of radar points, and subsequently transformed into the ego coordinate with the BEV augmentation matrix, as shown in Fig. \ref{fig:gaussian_coord_trans}. 
In addition, to fully leverage the camera modality, a Semantic Injection (SI) sub-module is introduced to provide semantic cues from image features for Gaussian attribute prediction. It not only enriches the Gaussian features, but also enables more accurate prediction of Gaussian poses and shapes.

The image branch follows LXL \cite{LXL} to adopt the sampling-based view transformation module. 
To save computation and reduce latency, the multi-level image features $\{\mathbf F_I^{(i)}\in\mathbb R^{C_i\times\frac{H_0}{2^i}\times\frac{W_0}{2^i}}\}_{i=0}^{L-1}$ from backbone \& neck are processed by an interpolation-based resolution alignment module to obtain a single feature map, eliminating the need for multi-scale sampling:
\begin{equation}
\begin{aligned}
    \mathbf F_{I,up}^{(i)}&=\mathtt{Upsample}_{H_0\times W_0}(\mathbf F_I^{(i)}),\,\,i=1,\cdots,L-1,\\
    \mathbf F_I&=\mathtt{Conv}_{1\times1}(\mathtt{Concat}(\mathbf F_I^{(0)},\{\mathbf F_{I,up}^{(i)}\}_{i=1}^{L-1})),
\end{aligned}
\end{equation}
where $\mathtt{Upsample}_{H\times W}(\cdot)$ is the bilinear interpolation operation with the target shape of $H_0\times W_0$, and $\mathbf F_I\in\mathbb R^{C\times H_0\times W_0}$ is the output single-scale image feature map.

For the fusion module, concatenation-based fusion is adopted for simplification. The detection head is the same as CenterPoint \cite{CenterPoint}, and the Box Gaussian Loss in RadarGaussianDet3D \cite{RadarGaussianDet3D} is added for bounding box attribute regression. 

\subsection{Ray-Centric Gaussian Primitives}

Inspired by GaussianBeV \cite{GaussianBeV} that predicts allocentric Gaussian rotation quaternions from image pixels, we predict ray-centric Gaussian primitives from radar points. Each Gaussian primitive is represented in a coordinate system related to the ray from radar origin to a radar point, namely ray-aligned coordinate system.

\begin{figure}
    \centering
    \includegraphics[width=\linewidth]{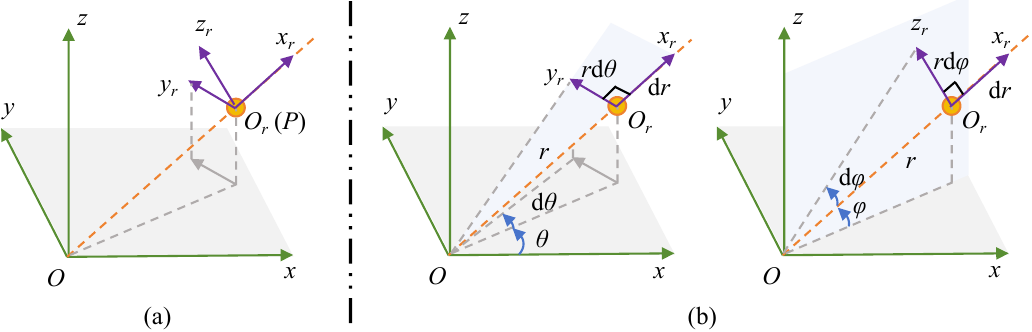}
    \caption{(a) The illustration of ray-aligned coordinate system $O_rx_ry_rz_r$. (b) When the range/azimuth/elevation of the radar point increases by a small value, the point will move along the $x_r$/$y_r$/$z_r$ axis. See Fig. \ref{fig:gaussian_coord_trans} for the legend.}
    \label{fig:ray-coord}
\end{figure}

\textbf{Ray-Aligned Coordinate System}. As illustrated in Fig. \ref{fig:ray-coord}(a), given a radar point $P$ in radar coordinate system $Oxyz$, we define ray-aligned coordinate system $O_rx_ry_rz_r$, where
\begin{itemize}
    \item $O_r$ coincides with the point $P$;
    \item $x_r$ is along the ray direction $\overrightarrow{OP}$;
    \item $y_r$ is parallel to plane $xOy$ and perpendicular to $x_r$, pointing left;
    \item $z_r$ is perpendicular to $x_r$ and $y_r$, pointing upwards.
\end{itemize}
In other words, if the Cartesian coordinate of a radar point is $\mathbf p_\textup{radar}=(x,y,z)^\top$, the direction vectors of axes $x_r,y_r,z_r$ in radar coordinate system ($\mathbf x_r,\mathbf y_r,\mathbf z_r$) can be calculated by
\begin{equation}
\begin{aligned}
    \mathbf x_r&=\mathbf p_\textup{radar}=(x,y,z)^\top,\\
    \mathbf y_r&=\mathbf z\times\mathbf x_r=(-y,x,0)^\top,\\\mathbf z_r&=\mathbf x_r\times\mathbf y_r=(-xz,-yz,x^2+y^2)^\top,
\end{aligned}
\end{equation}
where $\mathbf z=(0,0,1)^\top$ denotes the direction vector of $z$-axis. This definition is consistent with the range-azimuth-elevation measurement $(r,\theta,\varphi)$ of radar, where $r=\sqrt{x^2+y^2+z^2}, \theta=\arctan\frac yx,\varphi=\arcsin\frac zr$, as the direction of $x_r,y_r,z_r$ and $\textup dr,r\textup d\theta,r\textup d\phi$ are the same, as illustrated in Fig. \ref{fig:ray-coord}(b).

\textbf{Ray-Centric Gaussian Primitives}. For each radar point, the predicted Gaussian attributes, i.e., 3D mean offset $\Delta\boldsymbol{\mu}\in\mathbb R^3$, 3D scale $\mathbf s\in\mathbb R^3$, and rotation quaternion $\mathbf q\in\mathbb R^4$, are represented in the ray-aligned coordinate system. The covariance matrix can be calculated as 
\begin{equation}
    \boldsymbol{\Sigma}_\textup{ray}=\mathbf R\mathbf S\mathbf S^\top\mathbf R^\top,
\end{equation}
where $\mathbf S=\mathtt{diag}(\mathbf s)$, and $\mathbf R$ is the corresponding rotation matrix of quaternion $\mathbf q$.

To perform BEV rasterization, the ray-centric Gaussian primitives need to be transformed into the ego coordinate. 
As other attributes like opacity $o$ and feature $f^g$ are irrelevant to coordinate system, only positional offsets and covariance matrices are discussed in the following.

\textbf{Transform Gaussian Primitives to Radar Coordinate System}. The rotation matrix from radar coordinate system to the ray-aligned coordinate system can be obtained:
\begin{equation}
    \mathbf R_\textup{radar2ray}=\begin{bmatrix}
        \frac{\mathbf x_r}{\|\mathbf x_r\|_2}\\
        \frac{\mathbf y_r}{\|\mathbf y_r\|_2}\\
        \frac{\mathbf z_r}{\|\mathbf z_r\|_2}
    \end{bmatrix}=\begin{bmatrix}
        \frac xr & \frac yr & \frac zr\\
        -\frac y\rho & \frac x\rho & 0\\
        -\frac{xz}{r\rho} & -\frac{yz}{r\rho} & \frac{\rho}{r}
    \end{bmatrix},
\end{equation}
where $\rho=\sqrt{x^2+y^2}$. It can also be derived by calculating $\frac{[\partial r, r\partial\theta, r\partial\phi]}{\partial[x,y,z]}$. 

The covariance matrix and positional offset in radar coordinate can be calculated by
\begin{equation}
\begin{aligned}
    \boldsymbol{\Sigma}_\textup{radar}&=\mathbf R_\textup{radar2ray}^{-1}\boldsymbol{\Sigma}_\textup{ray}\mathbf R_\textup{radar2ray}^{-\top},\\
    \Delta\boldsymbol{\mu}_\textup{radar}&=\mathbf R_\textup{radar2ray}^{-1}\Delta\boldsymbol{\mu}.
\end{aligned}
\end{equation}

\begin{figure}
    \centering
    \includegraphics[width=\linewidth]{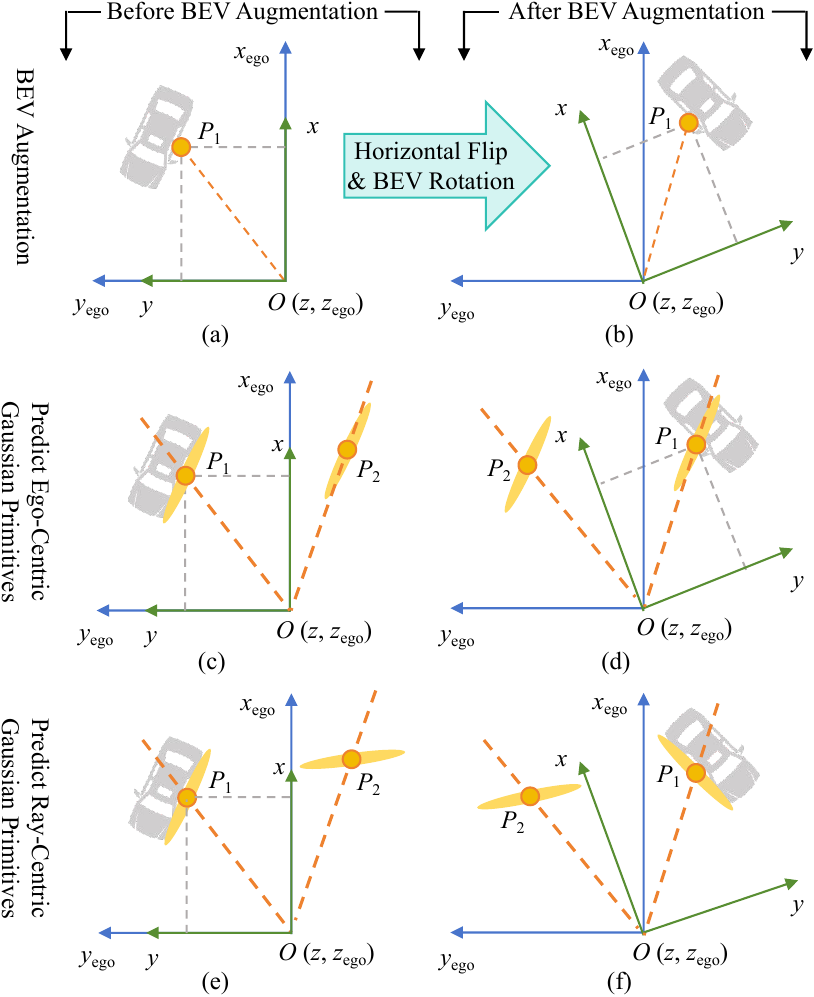}
    \caption{The comparison of ego-centric Gaussian primitives and ray-centric Gaussian primitives (top-down view). Position offsets are omitted. See Fig. \ref{fig:gaussian_coord_trans} for the legend.}
    \label{fig:ray_vs_ego}
\end{figure}

\textbf{Transform Gaussian Primitives to Ego Coordinate System}. In datasets like VoD\cite{VoD} and TJ4DRadSet \cite{TJ4DRadSet} that only have one front-facing radar, the ego coordinate system is usually defined coincide with the radar coordinate system (Fig. \ref{fig:ray_vs_ego}(a)). However, to facilitate training, BEV augmentation operations like rotation, scaling and horizontal flipping are usually conducted, making the radar coordinate system deviate from ego coordinate system, as shown in Fig. \ref{fig:ray_vs_ego}(b). As a result, transformation to ego coordinate is needed for Gaussian primitives before splatting to BEV. Giving the transformation matrix $\mathbf M_\textup{aug}$ introduced by BEV augmentation, where $\mathbf p_\textup{radar}=\mathbf M_\textup{aug}\mathbf p_\textup{ego}$, the covariance matrix and positional offset in ego coordinate can be calculated by
\begin{equation}
\begin{aligned}
    \boldsymbol{\Sigma}_\textup{ego}&=\mathbf M_\textup{aug}^{-1}\boldsymbol{\Sigma}_\textup{radar}\mathbf M_\textup{aug}^{-\top}\\
    &=(\mathbf R_\textup{radar2ray}\mathbf M_\textup{aug})^{-1}\boldsymbol{\Sigma}_\textup{ray}(\mathbf R_\textup{radar2ray}\mathbf M_\textup{aug})^{-\top},\\
    \Delta\boldsymbol{\mu}_\textup{ego}&=\mathbf M_\textup{aug}^{-1}\Delta\boldsymbol{\mu}_\textup{radar}=(\mathbf R_\textup{radar2ray}\mathbf M_\textup{aug})^{-1}\Delta\boldsymbol{\mu}.
\end{aligned}
\end{equation}

Finally, a Gaussian primitive in the ego coordinate can be expressed as $g=(\mathbf p_\textup{ego}+\Delta\boldsymbol{\mu}_\textup{ego},\boldsymbol{\Sigma}_\textup{ego},o,f^g)$. 

Fig. \ref{fig:ray_vs_ego}(c)-(f) compares ego-centric and ray-centric Gaussian primitives. Assuming that all points in the figure have the same feature vector, the predicted Gaussian attributes are identical in their represented coordinate system. As a result, ego-centric Gaussian primitives in Fig. \ref{fig:ray_vs_ego} (c) and (d) have the same position offset and orientation regardless of ray direction and radar pose, while ray-centric ones in Fig. \ref{fig:ray_vs_ego} (e) and (f) can align well with the object boundaries.

\subsection{Semantic Injection (SI)}
While the proposed R-PGE improves the geometric consistency of Gaussian primitives by predicting ray-centric Gaussian primitives, radar point clouds alone still lack semantic information, which is crucial for accurately predicting object boundaries and attributes. 
To mitigate this limitation and fully leverage the camera modality, we introduce the SI sub-module by incorporating visual features into the Gaussian attribute prediction process. 

Specifically, the SI sub-module is formulated as
\begin{equation}
\begin{aligned}
    f_I&=\mathtt{DeformAttn}(f_P,\mathbf F_I,\mathcal P(\mathbf p)),\\
    f&=\mathtt{Concat}(f_P,f_I),
\end{aligned}
\end{equation}
where $f_P$ and $\mathbf p$ are the feature and 3D coordinates of the radar point, and $f_I$ is the sampled image feature. $\mathcal{P}(\cdot)$ denotes the projection on the image plane, while $\mathtt{DeformAttn}(q,\mathbf F,\mathbf p_\textup{ref})$ is the deformable attention with query $q$, value $\mathbf F$ and reference point $\mathbf p_\textup{ref}$.
Note that this sub-module only introduces negligible latency, as the number of radar points is small.

The updated point feature $f$ is sent to an MLP to obtain attributes of ray-centric Gaussian primitives. 
Unlike conventional fusion strategies that merely mix features from different modalities, SI actively leverages semantics to refine both the appearance (i.e., features $f^g$) and the geometry (e.g., scales $\mathbf s$ and rotations $\mathbf q$) of Gaussian primitives, yielding BEV feature maps that better align with the underlying scene structure.

\section{Experiments}\label{sec:experiments}
\subsection{Implementation Details}
\textbf{Datasets and Evaluation Metrics.} In our experiment, the proposed model is evaluated on two publicly available multi-modal datasets: View-of-Delft \cite{VoD} (VoD) and TJ4DRadSet \cite{TJ4DRadSet}. Both datasets are collected with a 4D radar, a camera, and a LiDAR in driving scenes, containing temporally synchronized and spatially aligned data, as well as 3D bounding boxes of interested objects. The detection task on TJ4DRadSet is inherently more challenging than on VoD, as it encompasses a larger perception area, a broader set of object classes, and a wider variety of driving conditions. 

The official dataset splits and evaluation metrics are adopted in all experiments for fair comparison. Specifically, on the VoD dataset, the 3D Average Precisions (APs) over the entire annotated area (EAA) and the region of interest (ROI) are reported, where ROI is a small region in front of the ego vehicle. For the TJ4DRadSet, 3D AP and BEV AP are calculated. 

\textbf{Training Details}. Following \cite{RadarGaussianDet3D}, the training and benchmarking of the proposed model are both on a single NVIDIA Tesla V100 GPU. Multiple data augmentation are conducted during training, including BEV scaling, rotation and horizontal flipping, where calibration matrices are modified accordingly by calculating and multiplying $\mathbf M_\text{aug}$. The other training settings, e.g., the optimizer, learning rate, batch size, etc., are consistent with \cite{RadarGaussianDet3D}.

\subsection{Comparison with State-of-the-Arts}

\begin{table*}[]
\centering
\caption{Results on VoD \texttt{val} set}\label{tab:vod}
\begin{threeparttable}[b]
\begin{tabular}{cc|ccc|c|ccc|c|ccc}
\toprule
\multirow{2.5}{*}{Models} & \multirow{2.5}{*}{Modality} & \multicolumn{4}{c|}{EAA AP (\%)} & \multicolumn{4}{c|}{ROI AP (\%)} & \multicolumn{3}{c}{FPS (Hz)} \\ \cmidrule{3-13} 
 &  & Car & Ped. & Cyc. & mAP & Car & Ped. & Cyc. & mAP & V100 & 3090 & 4090 \\ \midrule
PointPillars (CVPR'19) \cite{PointPillars} & R & 38.8 & 34.4 & 66.9 & 46.7 & 71.9 & 45.1 & 88.4 & 67.8 & 78.4 & 178.4 & 187.0 \\
SMURF (T-IV'23) \cite{SMURF} & R & 42.3 & 39.1 & 71.5 & 51.0 & 71.7 & 50.5 & 86.9 & 69.7 & 30.0 & - & - \\ 
LGDD (IROS'25) \cite{LGDD} & R & 46.9 & 41.6 & 72.0 & 53.5 & 78.6 & 51.1 & 86.9 & 72.2 & - & 25.1 & - \\
SCKD$^\dagger$ (AAAI'25) \cite{SCKD} & R & 41.9 & 43.5 & 70.8 & 52.1 & 77.5 & 51.1 & 86.9 & 71.8 & - & - & 39.3 \\ 
MAFF-Net (RAL'25) \cite{MAFF-Net} & R & 42.3 & 46.8 & 74.7 & \textbf{54.6} & 72.3 & 57.8 & 87.4 & \textbf{72.5} & - & - & 28.7 \\
RadarGaussianDet3D (R-AL'26) \cite{RadarGaussianDet3D} & R & 40.7 & 42.4 & 73.0 & 52.0 & 71.2 & 51.7 & 89.0 & 70.6 & \textbf{83.2} & - & - \\ \midrule
RadarGaussianDet3D+ (\textbf{Ours}) & R & 39.9 & 45.9 & 75.0 & 53.6 & 71.0 & 52.3 & 89.0 & 70.8 & 77.5 & - & - \\ \midrule\midrule
RCFusion (T-IM'23) \cite{RCFusion} & R+C & 41.7 & 39.0 & 68.3 & 49.7 & 71.9 & 47.5 & 88.3 & 69.2 & - & 9.0 & - \\
LXL (T-IV'24) \cite{LXL} & R+C & 42.3 & 49.5 & 77.1 & 56.3 & 72.2 & 58.3 & 88.3 & 72.9 & 6.1 & - & - \\
LXLv2 (R-AL'25) \cite{LXLv2} & R+C & 47.8 & 49.3 & 77.2 & 58.1 & - & - & - & - & 6.5 & - & - \\
SGDet3D$^{\dagger\ddagger}$ (R-AL'25) \cite{SGDet3D} & R+C & 53.2 & 50.0 & 76.1 & 59.8 &  81.1 &  60.9 & 90.2 & 77.4 & - & - & 9.2 \\
HyDRa (ICRA'25) \cite{hydra} & R+C & 52.8 & 56.6 & 73.3 & 60.9 & 80.7 & 62.9 & 87.4 & 77.0 & - & - & - \\
RaGS$^{\dagger\ddagger}$ (ArXiv’25) \cite{RaGS} & R+C & 58.2 & 50.8 & 76.6 & 61.9 & 88.2 & 61.7 & 95.1 & 81.6 & - & - & 10.5 \\
MSSF-PP (T-ITS'25) \cite{MSSF} & R+C & 61.0 & 51.3 & 77.7 & 63.3 & 90.6 & 60.4 & 88.4 & 79.8 & - & 13.9 & - \\
CVFusion (ICCV'25) \cite{CVFusion} & R+C & 60.9 & 57.9 & 77.5 & 65.4 & 89.9 & \textbf{68.8} & 88.6 & 82.4 & - & 5.4 & - \\
SIFormer$^\ddagger$ (T-MM'26) \cite{SIFormer} & R+C & 53.8 & 50.7 & 76.1 & 60.2 & 81.2 & 60.5 & 90.2 & 77.3 & - & - & 6.9 \\
SIFormer$^{\dagger\ddagger}$ (T-MM'26) \cite{SIFormer} & R+C & \textbf{61.5} & 51.4 & 77.1 & 63.3 & \textbf{90.7} & 68.6 & 89.9 & 83.1 & - & - & 6.9 \\ \midrule
RCGDet3D (\textbf{Ours}) & R+C & 56.5 & \textbf{61.1} & \textbf{79.3} & \textbf{65.6} & 87.5 & 66.6 & \textbf{96.1} & \textbf{83.4} & \textbf{19.9} & - & - \\ \bottomrule
\end{tabular}
\begin{tablenotes}
    \item[1] In the column of Modality, R denotes radar, and C represents camera.
    \item[2] For clearer comparison, GPUs used for FPS measurement are specified, including Tesla V100, RTX 3090 and RTX 4090.
    \item[3] Models marked with $\dagger$ leverage LiDAR data during training (e.g., via distillation or depth supervision), while those marked with $\ddagger$ are trained with pseudo labels generated by Detectron2 \cite{detectron2}.
\end{tablenotes}
\end{threeparttable}
\end{table*}

\textbf{Results on VoD.} 
Table \ref{tab:vod} exhibits the results on the VoD dataset. 
It can be seen that the proposed RCGDet3D ranks first among 4D radar-camera fusion-based detectors in both EAA and ROI mAPs on VoD, even if some of the competitors leverage LiDAR data or pseudo labels during training. 
Moreover, RCGDet3D runs at 20 frames per second (FPS) on V100, while previous state-of-the-art methods barely reach 10 FPS even on the more powerful RTX 4090 hardware.
This significant speed advantage stems from our design choice to avoid the common computational bottleneck, i.e., multi-level image view transformation and sophisticated multi-modal fusion.
By focusing on improving the radar encoder via R-PGE and avoiding heavy view transformation, we achieve both superior accuracy and real-time inference. 

\textbf{Results on TJ4DRadSet.} 
As shown in Table \ref{tab:tj}, the results on TJ4DRadSet are in consistent with those on VoD, demonstrating the generalizability of our proposed RCGDet3D. 
Specifically, by only enhancing radar feature extraction, RCGDet3D outperforms the previous state-of-the-art models, whose high performance relies on their elaborate image view transformation and fusion modules. This comparison suggests that the quality of radar features plays a more critical role in radar-camera fusion.

\begin{figure}
    \centering
    \includegraphics[width=\linewidth]{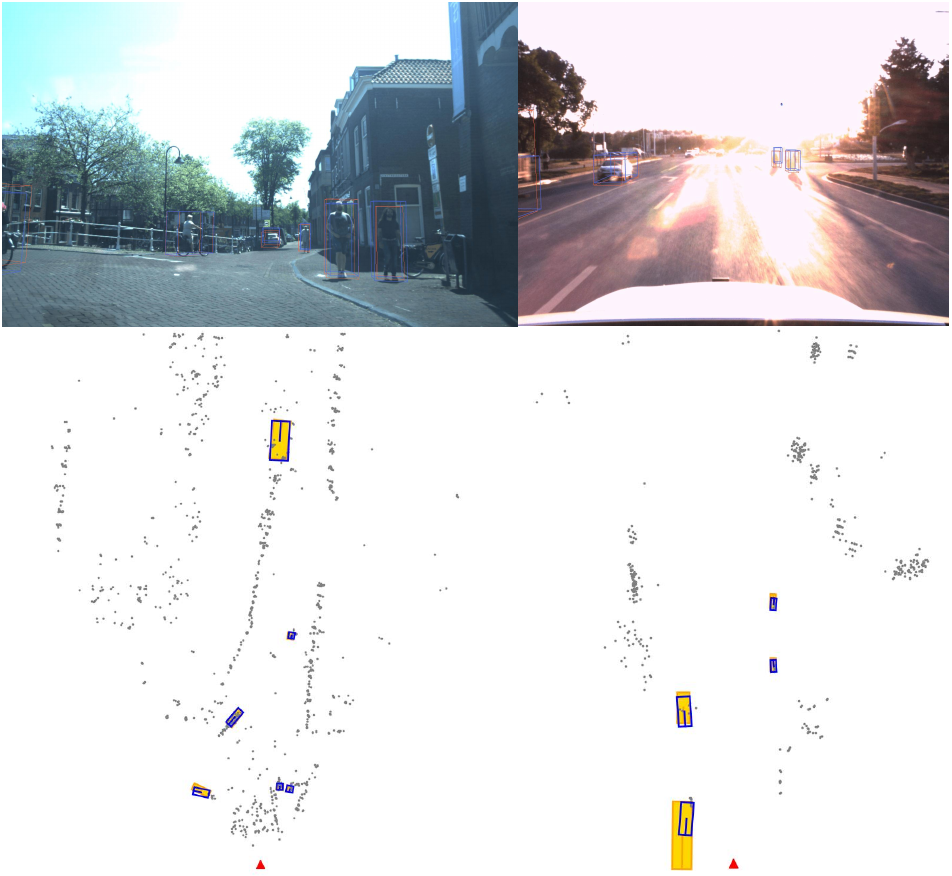}
    \caption{Visualization results of RCGDet3D on VoD \cite{VoD} (left) and TJ4DRadSet \cite{TJ4DRadSet} (right). Gray point, red triangle, orange rectangle and blue rectangle represent radar point, ego vehicle, ground-truth bounding box and predicted bounding box, respectively.}
    \label{fig:vis}
\end{figure}

Fig. \ref{fig:vis} visualizes the detection results of RCGDet3D.

\textbf{Comparisons of Radar Single Modality-based Detectors.}
Beyond multi-modal fusion, we also evaluate our radar-only variant RadarGaussianDet3D+, which is the enhanced version of RadarGaussianDet3D \cite{RadarGaussianDet3D} by replacing PGE with the proposed R-PGE (without SI).
It achieves detection accuracy on a par with state-of-the-art 4D radar-only detectors on both datasets, while maintaining high efficiency.
The performance gain demonstrates the effectiveness of R-PGE, setting a strong foundation for multi-modal fusion.

\begin{table*}[]
\centering
\caption{Results on TJ4DRadSet \texttt{test} set}\label{tab:tj}
\begin{threeparttable}[b]
\begin{tabular}{cc|cccc|c|cccc|c|c}
\toprule
\multirow{2.5}{*}{Models} & \multirow{2.5}{*}{Modality} & \multicolumn{5}{c|}{3D AP (\%)} & \multicolumn{5}{c|}{BEV AP (\%)} & \multirow{2.5}{*}{\makecell[c]{FPS\\(Hz)}} \\ \cmidrule{3-12} 
& & Car & Ped. & Cyc. & Tru. & mAP & Car & Ped. & Cyc. & Tru. & mAP & \\ \midrule
PointPillars (CVPR'19) \cite{PointPillars} & R & 21.26 & 28.33 & 52.47 & 11.18 & 28.31 & 38.34 & 32.26 & 56.11 & 18.19 & 36.23 & 42.9 \\
SMURF (T-IV'23) \cite{SMURF} & R  & 28.47 & 26.22 & 54.61 & 22.64 & 32.99 & 43.13 & 29.19 & 58.81 & 32.80 & 40.98 & 23.1\\
LGDD (IROS'25) \cite{LGDD} & R & 30.10 & 27.93 & 54.22 & 23.84 & 34.02 & 45.18 & 30.25 & 59.13 & 33.54 & 42.02 & - \\
MAFF-Net (R-AL'25) \cite{MAFF-Net} & R  & 27.31 & 33.13 & 54.35 & 26.71 & 35.38 & 39.05 & 35.25 & 56.35 & 35.73 & 41.59 & 17.9$^\diamond$ \\
RadarGaussianDet3D (R-AL'26) \cite{RadarGaussianDet3D} & R  & 26.69 & 28.18 & 65.84 & 19.63 & 35.08 & 41.66 & 30.28 & 69.62 & 26.35 & 41.98 & \textbf{43.5} \\ \midrule
RadarGaussianDet3D+ (\textbf{Ours}) & R  & 24.76 & 33.78 & 65.80 & 22.50 & \textbf{36.71} & 39.15 & 34.02 & 68.68 & 30.45 & \textbf{43.07} & 43.1 \\ \midrule\midrule
RCFusion (T-IM'23) \cite{RCFusion} & R+C  & 29.72 & 27.17 & 54.93 & 23.56 & 33.85 & 40.89 & 30.95 & 58.30 & 28.92 & 39.76 & 4.7$^\ast$ \\
LXL (T-IV'24) \cite{LXL} & R+C & - & - & - & - & 36.32 & - & - & - & - & 41.20 & - \\
LXLv2 (R-AL'25) \cite{LXLv2} & R+C & - & - & - & - & 37.32 & - & - & - & - & 42.35 & - \\
CVFusion (ICCV'25) \cite{CVFusion} & R+C & 51.54 & 29.49 & 49.41 & 29.55 & 40.00 & 58.07 & 31.65 & 51.29 & 35.29 & 44.07 & 5.7$^*$ \\
MSSF-PP (T-ITS'25) \cite{MSSF} & R+C & 52.04 & 35.11 & 55.72 & 24.14 & 41.75 & 64.31 & 38.39 & 60.08 & 30.86 & 48.41 & - \\ 
SGDet3D$^{\dagger\ddagger}$ (R-AL'25) \cite{SGDet3D} & R+C & 59.43 & 26.57 & 51.30 & 30.00 & 41.82 & 66.38 & 29.18 & 53.72 & 39.36 & 47.16 & - \\
RaGS$^{\dagger\ddagger}$ (ArXiv’25) \cite{RaGS} & R+C & - & - & - & - & 41.95 & - & - & - & - & 51.04 & - \\
SIFormer$^\ddagger$ (T-MM'26) \cite{SIFormer} & R+C & 61.12 & 27.18 & 46.11 & \textbf{38.22} & 43.15 & 68.30 & 31.47 & 46.46 & \textbf{45.64} & 47.96 & - \\ 
SIFormer$^{\dagger\ddagger}$ (T-MM'26) \cite{SIFormer} & R+C & \textbf{65.68} & 29.77 & 54.47 & 36.12 & 46.51 & \textbf{71.59} & 36.06 & 55.15 & 43.31 & 50.42 & - \\ \midrule 
RCGDet3D (\textbf{Ours}) & R+C & 56.96 & \textbf{36.29} & \textbf{70.51} & 26.88 & \textbf{47.66} & 69.33 & \textbf{38.81} & \textbf{73.30} & 34.88 & \textbf{54.08} & \textbf{11.6} \\ \bottomrule
\end{tabular}
\begin{tablenotes}
    \item[1] In the column of Modality, R denotes radar, and C represents camera.
    \item[2] For clearer comparison, GPUs used for FPS measurement are specified, including Tesla V100 (default), RTX 3090 ($\ast$) and RTX 4090 ($\diamond$).
    \item[3] Models marked with $\dagger$ leverage LiDAR data for depth supervision, while those marked with $\ddagger$ are trained with pseudo labels generated by Detectron2 \cite{detectron2}.
\end{tablenotes}
\end{threeparttable}
\end{table*}

\begin{table}[]
\centering
\caption{Ablation Studies of Components on VoD \texttt{val} set}\label{tab:ablation1}
\begin{tabular}{c|cc|c|cc|cc}
\toprule
\multirow{2.5}{*}{} & \multicolumn{2}{c|}{Gaussian Coordinates} & \multirow{2.5}{*}{Offsets} & \multicolumn{2}{c|}{SI} & \multicolumn{2}{c}{mAP (\%)} \\ \cmidrule{2-3}\cmidrule{5-8} 
 & Ego & Ray &  & BS & DA & EAA & ROI \\ \midrule
(a) & \Checkmark &  &  &  &  & 61.7 & 79.8 \\
(b) &  & \Checkmark &  &  &  & 62.5 & 80.7 \\
(c) &  & \Checkmark & \Checkmark &  &  & 64.9 & 81.1 \\
(d) &  & \Checkmark & \Checkmark & \Checkmark &  & 65.1 & 81.3 \\
(e) &  & \Checkmark & \Checkmark &  & \Checkmark & 65.6 & 83.4 \\\bottomrule
\end{tabular}
\end{table}

\subsection{Ablation Study}\label{sec:ablation}

\textbf{Ablations on Components.} 
To further validate the effectiveness of the proposed enhancements, ablation studies on VoD \cite{VoD} are conducted by progressively adding or replacing modules in the baseline detector. 
First, by changing the predicted Gaussian attributes from ego-centric to ray-centric, i.e., substituting PGE with R-PGE, the EAA mAP improves by 0.8\%. 
It validates that explicitly accounting for radar ray directions reduces the burden on the model to learn geometry from scratch, resulting in more effective feature extraction.
Subsequently, although RadarGaussianDet3D \cite{RadarGaussianDet3D} demonstrates the unnecessity of predicting positional offsets in PGE on TJ4DRadSet, we find that in R-PGE, they are of great help on VoD, bringing a performance gain of 2.4\% EAA mAP. 
This is possibly because the radar point clouds in VoD are the accumulation of multiple frames, where previous point clouds are transformed to align with the the current scan according to ego motion, neglecting the movement of dynamic objects. The predicted offsets can serve as motion compensation, moving the radar points to the correct position. 
On the other hand, as shown in Fig. \ref{fig:ray_vs_ego}(d), ego-centric Gaussian primitives have inaccurate orientations, adding positional offsets may increase this inaccuracy.
Finally, the SI sub-module further enhances the detection accuracy even if only bilinear sampling (BS) is adopted, demonstrating the importance of semantic cues during radar feature extraction. By applying the more powerful deformable attention (DA), the EAA mAP reaches 65.6\%.
 
\begin{figure}
    \centering
    \includegraphics[width=\linewidth]{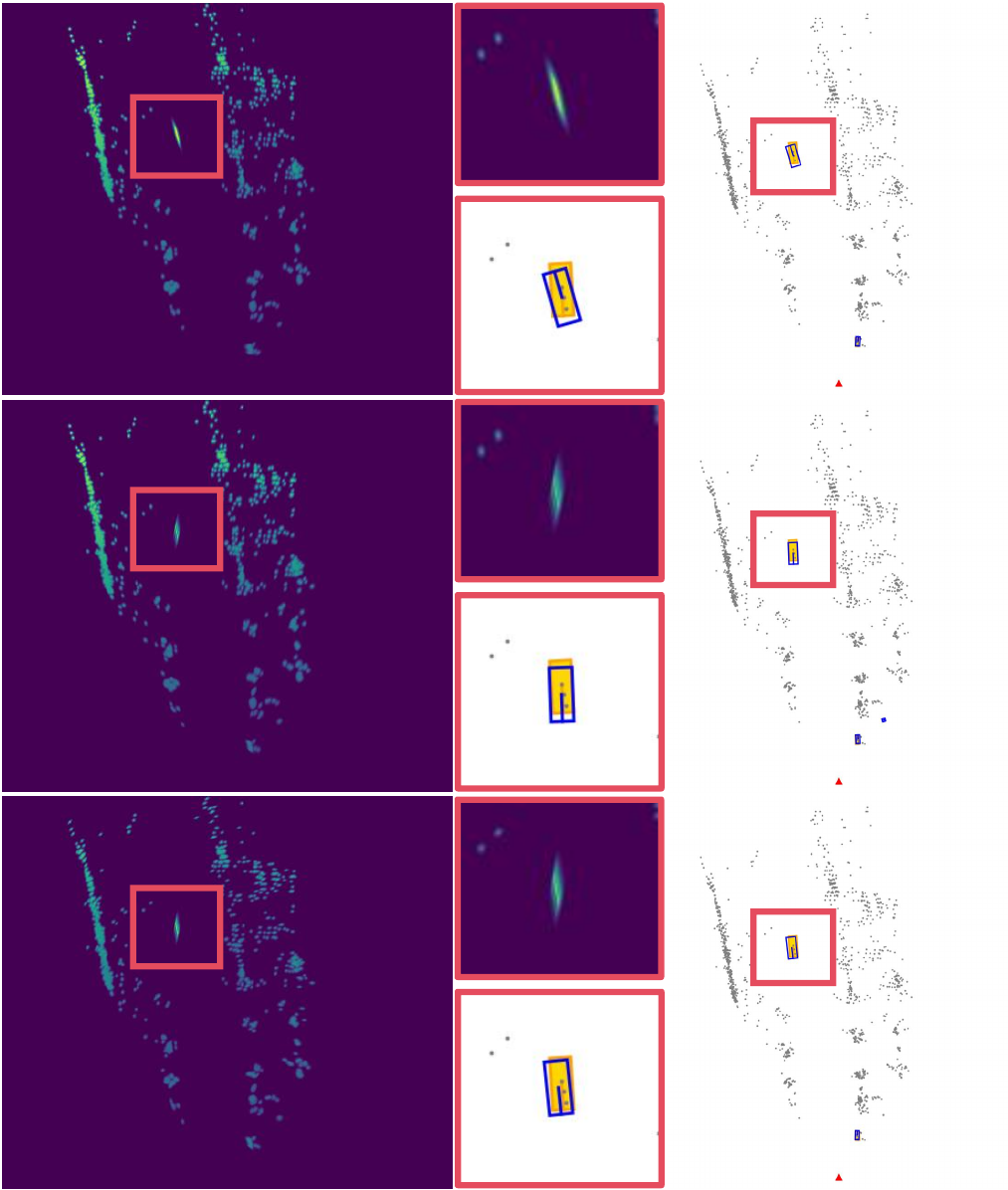}
    \caption{Visualization of BEV feature maps (left) from PGE (top), R-PGE (middle) and R-PGE with SI (bottom) on TJ4DRadSet \cite{TJ4DRadSet} \texttt{test} set. The right column displays the radar points and detection results in BEV.}
    \label{fig:bev_vis}
\end{figure}

\textbf{BEV Feature Map Visualization.} 
Fig. \ref{fig:bev_vis} visualizes the output BEV feature maps from PGE, R-PGE and R-PGE with SI. 
The visualizations are conducted on TJ4DRadSet \cite{TJ4DRadSet}, where our improvements yield more visually discernible differences on this dataset with denser radar point clouds. 
In the red box, the foreground region around a car is highlighted. 
It can be seen that due to the inaccurate Gaussian rotations in PGE, the detector predicts wrong orientation for the car. 
However, after replacing PGE with R-PGE, the Gaussian primitives are placed more reasonable, resulting in precise estimation for the car's yaw angle.
The phenomenon demonstrates that predicting ray-centric Gaussian primitives can facilitate learning Gaussian poses.
Moreover, when receiving semantic information via the SI sub-module, the background region becomes less activated, reducing the possibility of false positive detection.

\section{Conclusion}\label{sec:conclusion}
In this work, we uncover that the quality of radar feature extraction is the primary bottleneck in 4D radar-camera fusion-based 3D object detection, which has been largely overlooked by the community. 
Motivated by this finding, RCGDet3D is proposed, which shifts the focus from sophisticated fusion to improved radar feature encoding, resulting in accurate and fast detection. 
Its radar encoder is built upon the RadarGaussianDet3D \cite{RadarGaussianDet3D} framework with two key improvements. 
First, ray-centric Gaussian primitives instead of ego-centric ones are predicted, simplifying learning by eliminating the need to estimate the radar pose and ray direction, while also improving the geometric consistency.
Second, a Semantic Injection sub-module is introduced, where semantic cues from image features is integrated into the Gaussian attribute prediction process, 
resulting in more enriched radar feature maps.
Extensive experiments demonstrate that RCGDet3D and its radar-only variant both achieve state-of-the-art performance. 
Notably, by avoiding heavy computation on dense feature maps in image view transformation and multi-modal fusion, RCGDet3D is able to run significantly faster than previous methods, showcasing the potential of real-time deployment. 

By refocusing attention on the foundational role of radar feature quality, this work offers a new perspective for future research in multi-modal perception. Future research will focus on temporal modeling with the assistance of velocity information within the radar modality.

\bibliographystyle{IEEEtran}
\bibliography{reference}

@ARTICLE{LXL,
  author={Xiong, Weiyi and Liu, Jianan and Huang, Tao and Han, Qing-Long and Xia, Yuxuan and Zhu, Bing},
  journal={IEEE Transactions on Intelligent Vehicles}, 
  title={{LXL: LiDAR excluded lean 3D object detection with 4D imaging radar and camera fusion}}, 
  year={2024},
  volume={9},
  number={1},
  pages={79-92},
  doi={10.1109/TIV.2023.3321240}}

@inproceedings{RCBEVDet,
  title={{RCBEVDet: Radar-camera fusion in bird's eye view for 3D object detection}},
  author={Lin, Zhiwei and Liu, Zhe and Xia, Zhongyu and Wang, Xinhao and Wang, Yongtao and Qi, Shengxiang and Dong, Yang and Dong, Nan and Zhang, Le and Zhu, Ce},
  booktitle={Proceedings of the IEEE/CVF Conference on Computer Vision and Pattern Recognition},
  pages={14928--14937},
  year={2024}
}

@article{VoD,
  title={{Multi-class road user detection with 3+ 1D radar in the View-of-Delft dataset}},
  author={Palffy, Andras and Pool, Ewoud and Baratam, Srimannarayana and Kooij, Julian FP and Gavrila, Dariu M},
  journal={IEEE Robotics and Automation Letters},
  volume={7},
  number={2},
  pages={4961--4968},
  year={2022},
  publisher={IEEE}
}

@inproceedings{TJ4DRadSet,
  title={{TJ4DRadSet: A 4D radar dataset for autonomous driving}},
  author={Zheng, Lianqing and Ma, Zhixiong and Zhu, Xichan and Tan, Bin and Li, Sen and Long, Kai and Sun, Weiqi and Chen, Sihan and Zhang, Lu and Wan, Mengyue and others},
  booktitle={2022 IEEE 25th International Conference on Intelligent Transportation Systems (ITSC)},
  pages={493--498},
  year={2022},
  organization={IEEE}
}

@inproceedings{CenterPoint,
  title={{Center-based 3D object detection and tracking}},
  author={Yin, Tianwei and Zhou, Xingyi and Krahenbuhl, Philipp},
  booktitle={Proceedings of the IEEE/CVF conference on computer vision and pattern recognition},
  pages={11784--11793},
  year={2021}
}

@ARTICLE{SMURF,
  author={Liu, Jianan and Zhao, Qiuchi and Xiong, Weiyi and Huang, Tao and Han, Qing-Long and Zhu, Bing},
  journal={IEEE Transactions on Intelligent Vehicles}, 
  title={{SMURF: Spatial multi-representation fusion for 3D object detection with 4D imaging radar}}, 
  year={2024},
  volume={9},
  number={1},
  pages={799-812},
  doi={10.1109/TIV.2023.3322729}}

@article{RCFusion,
  title={{RCFusion: Fusing 4-D radar and camera with bird’s-eye view features for 3-D object detection}},
  author={Zheng, Lianqing and Li, Sen and Tan, Bin and Yang, Long and Chen, Sihan and Huang, Libo and Bai, Jie and Zhu, Xichan and Ma, Zhixiong},
  journal={IEEE Transactions on Instrumentation and Measurement},
  volume={72},
  pages={1--14},
  year={2023},
  publisher={IEEE}
}

@inproceedings{RPFA-Net,
  title={{RPFA-Net: A 4D radar pillar feature attention network for 3D object detection}},
  author={Xu, Baowei and Zhang, Xinyu and Wang, Li and Hu, Xiaomei and Li, Zhiwei and Pan, Shuyue and Li, Jun and Deng, Yongqiang},
  booktitle={2021 IEEE International Intelligent Transportation Systems Conference (ITSC)},
  pages={3061--3066},
  year={2021},
  organization={IEEE}
}

@inproceedings{PointPillars,
  title={{PointPillars: Fast encoders for object detection from point clouds}},
  author={Lang, Alex H and Vora, Sourabh and Caesar, Holger and Zhou, Lubing and Yang, Jiong and Beijbom, Oscar},
  booktitle={Proceedings of the IEEE/CVF Conference on Computer Vision and Pattern Recognition},
  pages={12697--12705},
  year={2019}
}

@inproceedings{PointNet,
  title={{PointNet: Deep learning on point sets for 3D classification and segmentation}},
  author={Qi, Charles R and Su, Hao and Mo, Kaichun and Guibas, Leonidas J},
  booktitle={Proceedings of the IEEE Conference on Computer Vision and Pattern Recognition},
  pages={652--660},
  year={2017}
}

@inproceedings{RadarPillars,
  title={{RadarPillars: Efficient Object Detection From 4D Radar Point Clouds}},
  author={Musiat, Alexander and Reichardt, Laurenz and Schulze, Michael and Wasenm{\"u}ller, Oliver},
  booktitle={2024 IEEE 27th International Conference on Intelligent Transportation Systems (ITSC)},
  pages={1656--1663},
  year={2024},
  organization={IEEE}
}

@ARTICLE{MAFF-Net,
  author={Bi, Xin and Weng, Caien and Tong, Panpan and Fan, Baojie and Eichberge, Arno},
  journal={IEEE Robotics and Automation Letters}, 
  title={MAFF-Net: Enhancing 3D Object Detection With 4D Radar via Multi-Assist Feature Fusion}, 
  year={2025},
  volume={10},
  number={5},
  pages={4284-4291}}

@article{3DGS,
  title={{3D Gaussian splatting for real-time radiance field rendering}},
  author={Kerbl, Bernhard and Kopanas, Georgios and Leimk{\"u}hler, Thomas and Drettakis, George},
  journal={ACM Trans. Graph.},
  volume={42},
  number={4},
  pages={139--1},
  year={2023}
}

@inproceedings{GaussianBeV,
  title={{GaussianBeV: 3D gaussian representation meets perception models for BeV segmentation}},
  author={Chabot, Florian and Granger, Nicolas and Lapouge, Guillaume},
  booktitle={2025 IEEE/CVF Winter Conference on Applications of Computer Vision (WACV)},
  pages={2250--2259},
  year={2025},
  organization={IEEE}
}

@inproceedings{GaussianLSS,
  title={Toward Real-world BEV Perception: Depth Uncertainty Estimation via Gaussian Splatting},
  author={Lu, Shu-Wei and Tsai, Yi-Hsuan and Chen, Yi-Ting},
  booktitle={Proceedings of the IEEE/CVF conference on Computer Vision and Pattern Recognition},
  pages={17124--17133},
  year={2025}
}

@inproceedings{GaussianFormer,
  title={{GaussianFormer: Scene as gaussians for vision-based 3D semantic occupancy prediction}},
  author={Huang, Yuanhui and Zheng, Wenzhao and Zhang, Yunpeng and Zhou, Jie and Lu, Jiwen},
  booktitle={European Conference on Computer Vision},
  pages={376--393},
  year={2024},
  organization={Springer}
}

@article{RaGS,
  title={{RaGS: Unleashing 3D Gaussian Splatting from 4D Radar and Monocular Cues for 3D Object Detection}},
  author={Bai, Xiaokai and Zhou, Chenxu and Zheng, Lianqing and Cao, Si-Yuan and Liu, Jianan and Zhang, Xiaohan and Zhang, Zhengzhuang and Shen, Hui-liang},
  journal={arXiv preprint arXiv:2507.19856},
  year={2025}
}

@inproceedings{SCKD,
  title={{SCKD: Semi-supervised cross-modality knowledge distillation for 4D radar object detection}},
  author={Xu, Ruoyu and Xiang, Zhiyu and Zhang, Chenwei and Zhong, Hanzhi and Zhao, Xijun and Dang, Ruina and Xu, Peng and Pu, Tianyu and Liu, Eryun},
  booktitle={Proceedings of the AAAI Conference on Artificial Intelligence},
  volume={39},
  number={9},
  pages={8933--8941},
  year={2025}
}

@article{LXLv2,
  title={{LXLv2: Enhanced LiDAR excluded lean 3D object detection with fusion of 4D radar and camera}},
  author={Xiong, Weiyi and Zou, Zean and Zhao, Qiuchi and He, Fengchun and Zhu, Bing},
  journal={IEEE Robotics and Automation Letters},
  year={2025},
  publisher={IEEE}
}

@article{SGDet3D,
  title={Sgdet3d: Semantics and geometry fusion for 3d object detection using 4d radar and camera},
  author={Bai, Xiaokai and Yu, Zhu and Zheng, Lianqing and Zhang, Xiaohan and Zhou, Zili and Zhang, Xue and Wang, Fang and Bai, Jie and Shen, Hui-Liang},
  journal={IEEE Robotics and Automation Letters},
  volume={10},
  number={1},
  pages={828--835},
  year={2024},
  publisher={IEEE}
}

@article{SIFormer,
  title={Boosting Instance Awareness via Cross-View Correlation with 4D Radar and Camera for 3D Object Detection},
  author={Bai, Xiaokai and Zheng, Lianqing and Cao, Si-Yuan and Zhang, Xiaohan and Wu, Zhe and Yu, Beinan and Wang, Fang and Bai, Jie and Shen, Hui-Liang},
  journal={arXiv preprint arXiv:2602.20632},
  year={2026}
}

@ARTICLE{MSSF,
  author={Liu, Hongsi and Liu, Jun and Jiang, Guangfeng and Jin, Xin},
  journal={IEEE Transactions on Intelligent Transportation Systems}, 
  title={MSSF: A 4D Radar and Camera Fusion Framework With Multi-Stage Sampling for 3D Object Detection in Autonomous Driving}, 
  year={2025},
  volume={26},
  number={6},
  pages={8641-8656}}

@inproceedings{LGDD,
  title={LGDD: Local-Global Synergistic Dual-Branch 3D Object Detection Using 4D Radar},
  author={Bai, Xiaokai and Yang, Qin and Zhou, Zili and Zhang, Fuyi and Wu, Zhe and Cao, Si-Yuan and Zheng, Lianqing and Yu, Beinan and Wang, Fang and Bai, Jie and others},
  booktitle={2025 IEEE/RSJ International Conference on Intelligent Robots and Systems (IROS)},
  pages={13318--13325},
  year={2025},
  organization={IEEE}
}

@inproceedings{SD4R,
  title={SD4R: Sparse-to-Dense Learning for 3D Object Detection with 4D Radar},
  author={Bai, Xiaokai and Cheng, Jiahao and Wang, Songkai and Luo, Yixuan and Zheng, Lianqing and Zhang, Xiaohan and Cao, Si-Yuan and Shen, Hui-Liang},
  booktitle={2025 IEEE 28th International Conference on Intelligent Transportation Systems (ITSC)},
  pages={4362--4368},
  year={2025},
  organization={IEEE}
}

@article{GSRender,
  title={Gsrender: Deduplicated occupancy prediction via weakly supervised 3d gaussian splatting},
  author={Sun, Qianpu and Shu, Changyong and Zhou, Sifan and Cheng, Runxi and Wei, Yongxian and Yu, Zichen and Yang, Dawei and Han, Sirui and Chun, Yuan},
  journal={arXiv preprint arXiv:2412.14579},
  year={2024}
}

@article{ODG,
  title={Odg: Occupancy prediction using dual gaussians},
  author={Shi, Yunxiao and Zhu, Yinhao and Han, Shizhong and Jeong, Jisoo and Ansari, Amin and Cai, Hong and Porikli, Fatih},
  journal={arXiv preprint arXiv:2506.09417},
  year={2025}
}

@inproceedings{hydra,
  title={Unleashing hydra: Hybrid fusion, depth consistency and radar for unified 3d perception},
  author={Wolters, Philipp and Gilg, Johannes and Teepe, Torben and Herzog, Fabian and Laouichi, Anouar and Hofmann, Martin and Rigoll, Gerhard},
  booktitle={2025 IEEE International Conference on Robotics and Automation (ICRA)},
  pages={7467--7474},
  year={2025},
  organization={IEEE}
}

@misc{detectron2,
  author={Yuxin Wu and Alexander Kirillov and Francisco Massa and Wan-Yen Lo and Ross Girshick},
  title={Detectron2},
  howpublished={\url{https://github.com/facebookresearch/detectron2}},
  year={2019}
}

@ARTICLE{RadarGaussianDet3D,
  author={Xiong, Weiyi and Zhu, Bing and Zheng, Zewei},
  journal={IEEE Robotics and Automation Letters}, 
  title={RadarGaussianDet3D: Gaussian Representation-Based Real-Time 3D Object Detection With 4D Automotive Radars}, 
  year={2026},
  volume={11},
  number={5},
  pages={5709-5716}}

@inproceedings{CVFusion,
  title={CVFusion: Cross-view fusion of 4D radar and camera for 3D object detection},
  author={Zhong, Hanzhi and Xiang, Zhiyu and Xu, Ruoyu and Fu, Jingyun and Xu, Peng and Wang, Shaohong and Yang, Zhihao and Pu, Tianyu and Liu, Eryun},
  booktitle={Proceedings of the IEEE/CVF International Conference on Computer Vision},
  pages={28188--28197},
  year={2025}
}
\end{document}